\documentclass[letterpaper]{article} 
\usepackage{aaai25}  
\usepackage{times}  
\usepackage{helvet}  
\usepackage{courier}  
\usepackage[hyphens]{url}  
\usepackage{graphicx} 
\usepackage{natbib}  
\usepackage{caption} 
\usepackage{algorithm}
\usepackage{algorithmic}
\usepackage{amsmath}
\usepackage{amssymb}
\usepackage{amsfonts}

\usepackage{newfloat}
\usepackage{listings}
\DeclareCaptionStyle{ruled}{labelfont=normalfont,labelsep=colon,strut=off} 
\floatstyle{ruled}
\newfloat{listing}{tb}{lst}{}
\floatname{listing}{Listing}
\title{Learning Fine-grained Domain Generalization via Hyperbolic State Space Hallucination}

\author{
    Qi Bi\textsuperscript{\rm 1},
    Jingjun Yi\textsuperscript{\rm 1},  
    Haolan Zhan\textsuperscript{\rm 2},
    Wei Ji\textsuperscript{\rm 3}\footnote{Corresponding authors}, 
    Gui-Song Xia\textsuperscript{\rm 1}\footnotemark[1]
    }
\affiliations{
    \textsuperscript{\rm 1}School of Artificial Intelligence, Wuhan University, Wuhan, China \\
    \textsuperscript{\rm 2}Faculty of Information Technology, Monash University, Melbourne, Australia \\
    \textsuperscript{\rm 3}School of Medicine, Yale University, New Haven, United States \\

    \{q\_bi, guisong.xia\}@whu.edu.cn, wei.ji@yale.edu
}

\usepackage{bibentry}

\begin{document}

\maketitle

\begin{abstract}
Fine-grained domain generalization (FGDG) aims to learn a fine-grained representation that can be well generalized to unseen target domains when only trained on the source domain data.
Compared with generic domain generalization, FGDG is particularly challenging in that the fine-grained category can be only discerned by some subtle and tiny patterns.
Such patterns are particularly fragile under the cross-domain style shifts caused by illumination, color and etc.
To push this frontier, this paper presents a novel Hyperbolic State Space Hallucination (HSSH) method.
It consists of two key components, namely, state space hallucination (SSH) and hyperbolic manifold consistency (HMC).
SSH enriches the style diversity for the state embeddings by firstly extrapolating and then hallucinating the source images.
Then, the pre- and post- style hallucinate state embeddings are projected into the hyperbolic manifold.
The hyperbolic state space models the high-order statistics, and allows a better discernment of the fine-grained patterns.
Finally, the hyperbolic distance is minimized, so that the impact of style variation on fine-grained patterns can be eliminated.
Experiments on three FGDG benchmarks demonstrate its state-of-the-art performance.
\end{abstract}

\begin{links}
\link{Code}{https://github.com/BiQiWHU/HSSH}
\end{links}

%

\section{Introduction}

Visual domain generalization is a fundamental task in both computer vision and machine learning, which handles the scenarios when the training samples and testing samples are not independently and identically distributed (i.i.d.).
It has been extensively studied in the past decade, which usually assumes that the domain gap is caused by the style variation, while the content (i.e., semantic category) is stable \cite{yan2020improve,chen2023domain,wang2023sharpness}.
Great success has been witnessed in advancing the generalization ability of a deep learning model in a variety of applications, such as semantic segmentation and object detection \cite{bi2024cmf,bi2024bwg,zhao2022style,lee2022wildnet,yi2024learning,bi2024fada}.
However, learning a domain generalized model for fine-grained visual categorization, termed as \textit{Fine-grained domain generalization} (FGDG), remains less explored.

\begin{figure}[!t]
    \centering
    \includegraphics[width=1.0\linewidth]{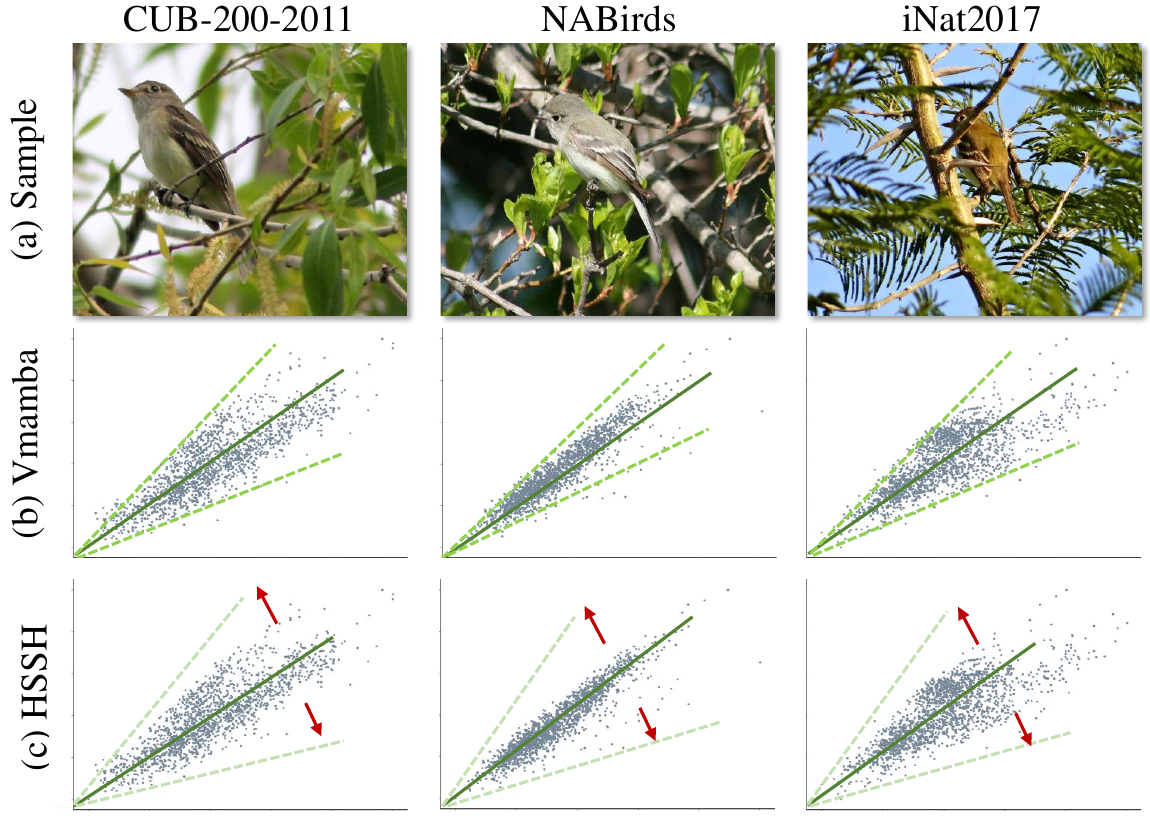}
    \vspace{-0.5cm}
    \caption{In each fine-grained domain (first row), due to the inter-category similarity, the embedding style from the VMamba baseline (second row) is limited. The proposed HSSH (third row) enriches the style diversity of the state embedding, which benefits the robustness to unseen domains. 
    The x- and y- axis refers to the mean and standard deviation value in the style space, ranging from 0 to 1.
    }
    \label{Fig1}
\end{figure}

FGDG holds both technical and practical significance. 
From a technical view, in generic domain generalization, the cross-domain content (i.e., semantic category) is stable and easy to differentiate. 
However, in FGDG, the fine-grained categories from a subordinate coarse-grained category can only be distinguished from some subtle and tiny patterns.
This problem becomes even more challenging when the cross-domain styles shift.
The shallower features of a deep model are sensitive to the style shift, posed by varied illumination, color, structure, etc.
The impact of such permutation accumulates from shallower to deeper features, making it more difficult to perceive the subtle fine-grained patterns.
From a practical view, FGDG also deserves further study.
Existing fine-grained visual categorization (FGVC) datasets are still small in amount (10K). A pre-trained FGVC model is likely to encounter unseen and out-of-domain samples when doing real-world inference. 

The emerging selective state space model (SSM) \cite{gu2021combining,gu2021efficiently}, exemplified by the VMamba \cite{zhu2024vision} and Vision Mamba \cite{liu2024vmamba}, has become a recent paradigm.
It is able to exploit the relation between local- and global- representations by selective scan mechanism and recurrent modeling from the patch embeddings of an image.
This property makes SSM have great potential for FGDG, as the local subtle patterns can be handled by the selective scan mechanism and patch embedding, while the scene content can be handled by the recurrent modeling to exploit the long-range dependency.  

\begin{figure}[!t]
    \centering
    \includegraphics[width=1.0\linewidth]{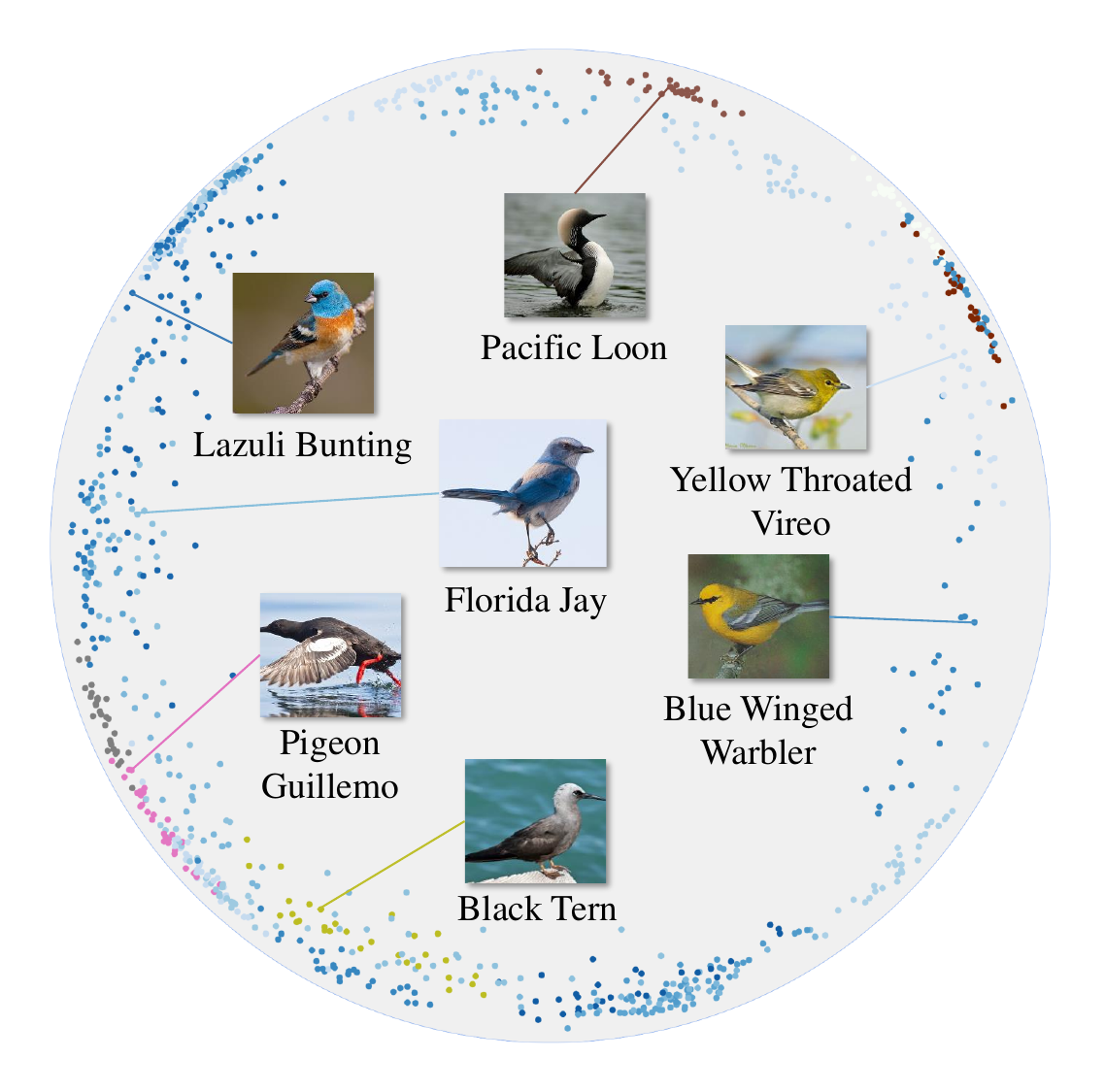}
    \vspace{-0.9cm}
    \caption{Hyperbolic manifold provides a feasible path to learn hierarchical and high-order statistics, which has potential to distinguish a fine-grained category from other fine-grained categories under the same coarse-grained category. 
    }
    \label{Fig2}
\end{figure}

In this paper, we aim to advance SSM for FGDG, by allowing it to simultaneously improve the robustness to cross-domain style shift and exploit the fine-grained semantic category representation.
To realize the first objective, we propose a state space hallucination (SSH) technique, which hallucinates the state embeddings from each VMamba layer. 
By implementing linear extrapolation on the per-batch state embeddings, it enriches the style diversity than the VMamba baseline (shown in Fig.~\ref{Fig1}).
The enriched styles allow the state embeddings to be less sensitive to the domain shift when inference on unseen target domains, benefiting the learning of subtle fine-grained patterns.

To realize the second objective, a hyperbolic manifold consistency (HMC) learning scheme is proposed.
Specifically, both pre- and post- hallucinated state embeddings are projected into the hyperbolic manifold, which is well recognized to preserve the hierarchical and high-order statistics \cite{ganea2018hyperbolic, yu2019numerically}.
The hyperbolic manifold allows the state embeddings from the same fine-grained semantic category to be as close as possible, while the state embeddings from different fine-grained categories to be as apart as possible (shown in Fig.~\ref{Fig2}).
Afterwards, the hyperbolic distance between the pre- and post- hallucinated state embedding is minimized, so as to alleviate the impact from cross-domain styles on differentiating the fine-grained semantic category.

The proposed hyperbolic state space hallucination (HSSH) method is empowered by the above two key components, namely, SSH and HMC.
Notably, it not only makes an early exploration of how to harness SSM for FGDG, but also of how to embed the hyperbolic manifold into the state space.
The proposed HSSH is validated on three FGDG benchmarks.
In each FGDG benchmark, two or more fine-grained visual categorization (FGVC) datasets from the same coarse-grained category (e.g., bird) are used as the source and target domain, and the common fine-grained categories among these datasets are selected.

Our contributions can be summarized as follows.
\begin{itemize}
    \item We initiate an early exploration to harness State Space Models (SSM) for fine-grained domain generalization. 
    \item A Hyperbolic State Space Hallucination (HSSH) method is proposed, which innovatively embeds the hyperbolic manifold into the selective state space.
    \item We propose a state space hallucination (SSH) component to enrich the style diversity for SSM, and a hyperbolic manifold consistency (HMC) component to learn discriminative fine-grained categories under domain shift.
    \item Extensive experiments show that the proposed HSSH significantly outperforms the state-of-the-art by up to 16.14\%, 26.01\% and 12.54\% on CUB-Paintings, RS-FGDG and Birds-31, respectively.
\end{itemize}

\section{Related Work}

\textbf{Fine-grained Domain Generalization}
Over the past decade, Fine-grained Visual Categorization (FGVC) have been extensively studied \cite{zhao2021part,hu2021rams,du2021progressive,yang2022fine}.
However, these methods assume that the training and testing images are i.i.d., which is far from the reality.
FGDG handles this challenge.
Specifically, a progressive learning scheme is proposed to maintain the consistency of fine-grained semantics for FGVC on different domains \cite{du2021progressive}. 
A progressive adversarial network (PAN) along with a curriculum-based adversarial learning scheme is proposed to align fine-grained categories across domains  \cite{wang2021dynamic}.
More recently, a feature structuralization method is proposed to stabilize the fine-grained semantics across domains \cite{yu2024fine}.


\noindent \textbf{Style Hallucination} 
has been recognized as a common solution to enrich style diversity \cite{huang2017arbitrary,chen2021hinet,zhang2022exact}.
It has also been widely explored in domain generalized semantic segmentation  \cite{zhao2022style,zhong2022adversarial,zhao2024style,yi2024hallucinated}, where the augmented styles enhance the generalization to unseen domains that have different styles. 

\begin{figure*}[!t]
    \centering
    \includegraphics[width=1.0\linewidth]{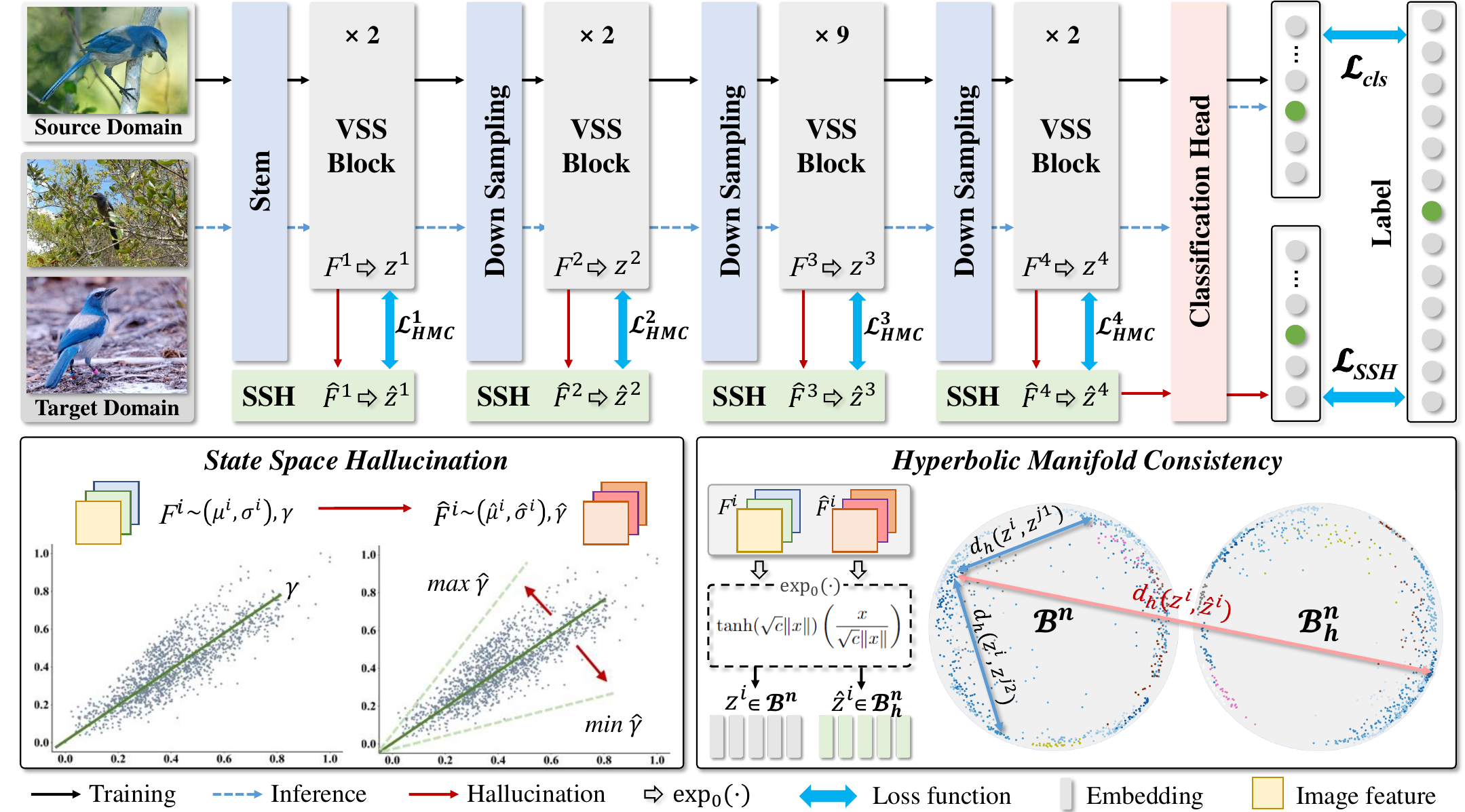}
    \vspace{-0.5cm}
    \caption{The proposed hyperbolic state space hallucination (HSSH) is empowered by two key components, namely, state space hallucination (SSH) and hyperbolic manifold consistency (HMC). SSH enriches the style diversity for the state embeddings by firstly extrapolating and then hallucinating from the source images. HMC minimizes the hyperbolic distance between the pre- and post- hallucinated samples, so that the impact from the style variation on fine-grained patterns can be eliminated.}
    \label{Fig3}
\end{figure*}

\noindent \textbf{State Space Models}
(SSM) \cite{kalman1960new} is fundamental for various fields, and has recently shown great success for deep representation learning \cite{gu2021combining,gu2021efficiently,gu2023mamba,bi2024samba}. 
This has led to the development of Mamba variants such as Vision Mamba \cite{liu2024vmamba} and VMamba \cite{zhu2024vision} for computer vision.
Nevertheless, the generalization ability of SSM on unseen target domains, especially for fine-grained semantic categories, remains to be explored. 

\noindent \textbf{Hyperbolic Manifold Learning}
has garnered significant interest  \cite{ganea2018hyperbolic, yu2019numerically} and have been applied to many computer vision tasks \cite{yang2023hyperbolic,pmlr-v202-desai23a,hu2024rethinking,kong2024hyperbolic,mettes2024hyperbolic,xu2023hyperbolic,weber2020robust,pmlr-v202-desai23a,hu2024rethinking,kong2024hyperbolic}. 
However, how to embed hyperbolic alignment within state space or for FGDG remains an open question.

\section{Methodology}

\subsection{Preliminaries \& Overview}

Given $K$ unseen target domains $\mathcal{D}_1, \dots, \mathcal{D}_K$ and a single source domain $\mathcal{D}_{K+1}$, we denote the joint image and label set within domain $\mathcal{D}_k$ as $\{(x_n^k, y_{c,n}^k, y_{f,n}^k)\}_{n=1}^{N_k}$, where $k = 1, 2, \dots, K$, and $N_k$ represents the number of samples in domain $\mathcal{D}_k$. Here, $y_{c,n}^k$ and $y_{f,n}^k$ denote the coarse-grained and fine-grained label of the sample $x_n^k$, respectively. 
Each sample among these domains shares a common coarse-grained category $y_{c,n}^k$ but varies in different fine-grained categories $y_{f,n}^k$.
Our objective is to learn a model $f_{\theta}: x_n^k \rightarrow y_{f,n}^k$, trained solely on the single source domain $\mathcal{D}_{K+1}$, that generalizes to the unseen target domains.

Fig.~\ref{Fig3} gives an overview of the proposed Hyperbolic State Space Hallucination (HSSH) scheme. 
On top of the VMamba backbone, it consists of two key components, namely, State Space Hallucination (SSH), and Hyperbolic Manifold Consistency (HMC).
SSH enriches the style space $\mathbb{S}$ and then hallucinates the state embeddings of samples $x_n^{k+1}$ from the source domain $\mathcal{D}_{K+1}$. 
HMC projects the pre- and post- hallucinated state embeddings into the hyperbolic manifolds, 
where they are aligned by minimizing the geodesic distance $d$ between hyperbolic embeddings $\mathbf{z}$. It helps promote stable representation of the fine-grained semantic separability in the state space.

\subsection{State Space Hallucination}

For FGVC, the key challenge is to identify the subtle and tiny fine-grained patterns that represent the fine-grained semantics from numerous patches of an image. 
The Mamba architecture \cite{gu2023mamba} encodes the input data into sequences, implements the selective scan mechanism and recurrent modeling \cite{gu2021efficiently} to highlight the relevant information \cite{lu2024structured}. 
This property has great potential to discern the key patterns, but the robustness to style variation needs to be enhanced for cross-domain scenario.
The state space hallucination (SSH) component is proposed to push this frontier. 

The Vision Mamba encoder \cite{zhu2024vision} consists of four blocks from shallow to deep. 
Assume the state embeddings from the $i$-th block are denoted as $\mathbf{F}^{i} \in \mathbb{R}^{B \times C_{i} \times H_{i} \times W_{i}}$, where $C_{i}, H_{i}, W_{i}$ denote the channel, height and width of a patch, respectively. 
Following the existing definition \cite{huang2017arbitrary,chen2021hinet,zhang2022exact}, the style of the state embeddings is quantified by computing the channel-wise mean $\mu$ and standard deviation $\sigma$.
Specifically, given a batch of samples from the source domain $\{x_{n}\}_{n=1}^{B}$, where $B$ denotes the batch size, the channel-wise mean $\boldsymbol \mu^{i} \in \mathbb{R}^{B \times C_{i}}$ and deviation $\boldsymbol \sigma^{i} \in \mathbb{R}^{B \times C_{i}}$ can be mathematically computed as:
\begin{gather}
\boldsymbol \mu^{i} = \frac{1}{H_i W_i} \mathbf{F}^{i}, \notag \\
\boldsymbol \sigma^{i} = \sqrt{\frac{1}{H_i W_i} \sum_{h, w \in H_i, W_i} (\mathbf{F}^{i} - \boldsymbol \mu^{i})(\mathbf{F}^{i} - \boldsymbol \mu^{i})^\mathrm{T}}.
\label{eq3}
\end{gather}

As shown in Fig.~\ref{Fig1}b, the style distribution of the state embeddings is highly constrained in the style space $\mathcal{S} \subset \mathbb{R}^{[0,1] \times [0,1]}$, where the upper and lower slopes (green lines) are relatively close to each other. 
Such limited diversity makes it less generalized to unseen target domains with different styles.
To handle this, we enrich its diversity by style hallucination, which enlarges the differences between the upper and lower slopes (in Fig.~\ref{Fig1}c).
The general idea is to introduce another slope in $\mathcal{S}$ to represent the hallucinated styles without prior knowledge of other domains.
Considering $(\boldsymbol \mu_{j, k}^{i}, \boldsymbol \sigma_{j, k}^{i})$ as samples points, where $0 \leq j \leq B, 0 \leq k \leq C_{i}$, we approximate the slope value $\boldsymbol \gamma^{i}$ from the samples in a batch $\{x_{n}\}_{n=1}^{B}$ by the least square method:
\begin{equation}
\boldsymbol \gamma^{i} = \frac{\sum_{j=1}^{B} \sum_{k=1}^{C_{i}} (\boldsymbol \mu_{j, k}^{i} - \bar{\boldsymbol \mu}^{i})(\boldsymbol \sigma_{j, k}^{i} - \bar{\boldsymbol \sigma}^{i})}{\sum_{j=1}^{B} \sum_{k=1}^{C_{i}} (\boldsymbol \mu_{j, k}^{i} - \bar{\boldsymbol \mu}^{i})^2},
\label{eq4}
\end{equation}
where $\bar{\boldsymbol \mu}^{i}$ and $\bar{\boldsymbol \sigma}^{i}$ denote the average values of $\mu^{i}$ and $\boldsymbol \sigma^{i}$. 

Then, the style range is expanded by altering the slope value and sampling style points on the line of the new slope.
We calculate the maximum slope $\max{\boldsymbol \gamma^{i}}$ and the minimum slope $\min{\boldsymbol \gamma^{i}}$ of the original state embeddings. Then, we sample new points on the expanded style regions with a broader range of slope values ($\min \widetilde{\boldsymbol \gamma} \leq \widetilde{\boldsymbol \gamma} \leq \max \widetilde{\boldsymbol \gamma}$) by
$\min \widetilde{\boldsymbol \gamma} = 2\min \boldsymbol \gamma - \max \boldsymbol \gamma$ and 
$\max \widetilde{\boldsymbol \gamma} = 2\max \boldsymbol \gamma - \min \boldsymbol \gamma$.
Here, the new style slopes are sampled with random slope $\widetilde{\boldsymbol \gamma} \sim [\min \widetilde{\boldsymbol \gamma}, \max \widetilde{\boldsymbol \gamma}]$. 
Then we obtain hallucinated features $\widetilde{\mathbf{F}}^{i}$ from sampled points ($\widetilde{\boldsymbol \mu}^{i}, \widetilde{\boldsymbol \sigma}^{i}$), given by:
\begin{equation}
\widetilde{\mathbf{F}}^{i} = \widetilde{\boldsymbol \sigma}^{i} \frac{\mathbf{F}^{i}-\boldsymbol \mu^{i}}{\boldsymbol \sigma^{i}} + \widetilde{\boldsymbol \mu}^{i}.
\label{eq6}   
\end{equation}


\subsection{Hyperbolic Manifold Consistency}

The enrichment of style diversity does not necessarily ensure that the state space can precisely discern the subtle and tiny fine-grained patterns from a number of image patches.
To address this issue, we further project the state embeddings into the hyperbolic manifold, which is well recognized to represent the hierarchical and high-order information \cite{ganea2018hyperbolic, yu2019numerically}.
This property allows the samples from different fine-grained categories to fall apart.
We propose Hyperbolic Manifold Consistency (HMC) to constrain the pre- and post- hallucinated style embeddings, so that the fine-grained representation can be consistent despite the style shift. 

Specifically, let the hallucinated features from the four encoder blocks are denoted as $\widetilde{\mathbf{F}}^{1}$, $\widetilde{\mathbf{F}}^{2}$, $\widetilde{\mathbf{F}}^{3}$ and $\widetilde{\mathbf{F}}^{4}$, respectively. 
On the other hand, Poincar\'e ball is selected to as the consistency manifold since it has been widely utilized in computer vision community. The Poincaré ball model is characterized by the pair \((\mathcal{B}^n_c, g_c^{\mathcal{B}})\), where the manifold is defined as \(\mathcal{B}^n_c = \{ x \in \mathbb{R}^n : c\|x\|^2 < 1\}\). 
As the fine-grained patterns exist in the Euclidean space,
a transformation is needed to work within the hyperbolic space of the Poincaré ball.
This objective is achieved by mapping the state embedding $\mathbf{F}^{i}$ using the exponential map centered at a reference point \(v\):
\begin{equation}
    \exp_v^c(\mathbf{F}^{i}) = v \oplus_c \left( \tanh\left(\frac{\sqrt{c}\| \mathbf{F}^{i} \|}{2}\right) \frac{ \mathbf{F}^{i} }{\sqrt{c}\| \mathbf{F}^{i} \|} \right),
    \label{eq8}   
\end{equation}
where \(\oplus_c\) denotes the Möbius addition, given by
\begin{equation}
    v \oplus_c w = \frac{(1 + 2c\langle v, w \rangle + c\|w\|^2)v + (1 - c\|v\|^2)w}{1 + 2c\langle v, w \rangle + c^2\|v\|^2\|w\|^2}.
    \label{eq9}   
\end{equation}

Practically, the point \(v\) is often taken as the origin. The exponential map expression can be simplified as
\begin{equation}
    \exp_0(\mathbf{F}^{i}) = \tanh(\sqrt{c}\|\mathbf{F}^{i} \|)\left(\frac{\mathbf{F}^{i}}{\sqrt{c}\|\mathbf{F}^{i}\|}\right).
    \label{eq10}   
\end{equation}

Let $\boldsymbol{z}^{i} \in \mathbb{R}^{B \times C \times H_i \times W_i}, \widetilde{\boldsymbol{z}}^{i} \in \mathbb{R}^{B \times C \times H_i \times W_i}$ denote the hyperbolic embedding of the pre-hallucinated state embedding $\mathbf{F}^{i}$ and the post-hallucinated state embedding $\widetilde{\mathbf{F}}^{i}$, the projection to the Poincar\'e hyperbole (by Eq.~\ref{eq10}) can be re-formulated as $\boldsymbol z^{i} = \exp_0(\mathbf{F}^{i}), \widetilde{\boldsymbol z}^{i} = \exp_0(\widetilde{\mathbf{F}}^{i})$.
Then, the hyperbolic distance $d_h$ can be calculated as
\begin{equation}
    d_h(\boldsymbol z^{i}, \widetilde{\boldsymbol z}^{i}) = \frac{2}{\sqrt{c}} \tanh^{-1} \left( \sqrt{c} \left\| -\boldsymbol z^{i} \oplus_c \widetilde{\boldsymbol z}^{i} \right\| \right),
    \label{eq12}   
\end{equation}
where $c$ is a hyperparameter that controls the radius of the hyperbolic space, and by default it is set to 0.1.  
To ensure that the fine-grained representation is consistent between the pre- and post- hallucinated hyperbolic embeddings, we devise a hyperbolic manifold consistency loss:
\begin{equation}
\mathcal{L}_{HMC} = \left( d_h(\boldsymbol z^{i}, \widetilde{\boldsymbol z}^{i}) + \frac{1}{N_{S}} \sum_{(i, j) \in S} d_h(\boldsymbol z^{i}, \boldsymbol z^{j}) \right),
\label{eq13}
\end{equation}
where $S$ denotes the set of sample pairs $(\boldsymbol z_i, \boldsymbol z_j)$ that belong to the same coarse-grained category, and $N_{S}$ denotes the number of these pairs.

\subsection{Implementation Details}

The pre- and post- hallucinated state embeddings $\mathbf{F}_{c}^{4}$ and $\widetilde{\mathbf{F}}_{c}^{4}$ from the last block of the VMamba encoder are fed into the classifier $\phi (\cdot)$, optimized by the classification loss function $\mathcal{L}_{cls}$ and $\widetilde{\mathcal{L}}_{cls}$, respectively.
$\phi (\cdot)$ is a linear layer with an input/output dimension of 768/category number, respectively.
Assume that \( C \) represents the category number. Then, the classification loss is defined as
\begin{gather}
\mathcal{L}_{cls} = -\frac{1}{B} \sum_{i=1}^{B} \sum_{j=1}^{C} y_{i,j} \log(\phi(\mathbf{F}^4)_{i,j}).
\label{eq14}
\end{gather}

The total loss $\mathcal{L}$ is a linear combination between the classification loss and the hyperbolic manifold consistency $\mathcal{L}_{HMC}$, given by
\begin{equation}
\mathcal{L} = \mathcal{L}_{cls} + \widetilde{\mathcal{L}}_{cls} + \lambda \mathcal{L}_{HMC},
\label{eq15}
\end{equation}
where $\lambda$ is a hyper-parameter used to balance semantic consistency in the state space and hierarchical consistency in the hyperbolic space. Empirically, $\lambda$ is set to 0.5.

For all experiments across the three FGDG settings, the Adam optimizer is employed with a learning rate of $1 \times 10^{-4}$, and momentum parameters set to 0.9 and 0.99. The training process spans 100 epochs. 

\section{Experiments}
\subsection{Dataset and Evaluation Protocols}

\textbf{CUB-Paintings} encompasses two domains: Caltech-UCSD Birds-200-2011 (CUB-200-2011, denoted as C) \cite{wah2011caltech} and CUB-200-Paintings (CUB-P, denoted as P) \cite{wang2020progressive}. 
CUB-200-2011 contains 11,788 images across 200 bird subcategories. 
CUB-P shares the same categorization system and includes 3,047 images featuring artistic representations such as watercolors, oil paintings, pencil drawings, stamps, and cartoons. 
These datasets exhibit notable domain shifts due to different styles. 
Furthermore, both domains can be grouped into 38 coarse categories.

\noindent \textbf{RS-FGDG} comprises two domains: Million-AID \cite{long2021creating} (MAID, denoted as M) and NWPU-RESISC45 \cite{cheng2017remote} (NWPU, denoted as N). 
Million-AID contains 1,000,848 scenes organized into 51 fine-grained categories. 
NWPU includes 31,500 images across 45 scene classes, each with 700 images. 
For domain generalization, the 22 common fine-grained categories from 8 coarse-grained categories are used.

\noindent \textbf{Birds-31} comprises three domains: Caltech-UCSD Birds-200-2011 (CUB-200-2011, denoted as C) \cite{wah2011caltech}, NABirds (denoted as N) \cite{van2015building}, and iNaturalist2017 (iNat2017, denoted as I) \cite{van2018inaturalist}. 
NABirds consists of 555 categories, with a total of 48,562 images. 
iNat2017 contains 859,000 images spanning over 5,000 species of plants and animals.
For FGDG, 31 common categories were selected, resulting in a total of 7,693 images, with 1,848 from CUB-200-2011, 2,988 from NABirds, and 2,857 from iNat2017. These fine-grained categories are consolidated into four coarse-grained categories.

In line with previous studies \cite{yang2022fine, Hsu2023ABC, Song2023SEB}, top-1 accuracy is used as the evaluation metric on unseen target domains. 

\begin{table}[!t]
	\centering
	\resizebox{\linewidth}{!}{
	\begin{tabular}{c|ccc|c}
		\hline
		Method & C $\rightarrow$ P &  P $\rightarrow$ C & Avg & Params\\ 
        \cline{1-5}
        \textit{ResNet-Based:} & \\
        DP-Net* (DA) & 65.22 & 47.81 & 56.52 & 103M \\
        PAN* (DA) & 67.40 & 50.92 & 59.16 & 103M \\
        ARM & 47.98 & 31.53 & 39.76 & 24M \\
        DANN & 54.06 & 37.09 & 45.57 & 24M \\
        MLDG & 55.40 & 34.15 & 44.78 & 23M \\
        GroupDRO & 54.94 & 35.67 & 45.31 & 23M \\
        CORAL & 54.70 & 35.29 & 45.00 & 23M \\
        SagNet & 56.33 & 36.71 & 46.52 & 24M \\
        MixStyle & 52.97 & 28.44 & 40.71 & 23M \\
        Mixup & 54.58 & 34.66 & 44.62 & 23M \\
        RIDG & 36.41 & 24.11 & 30.26 & 24M \\
        SAGM & 57.83 & 37.16 & 47.50 & 23M \\
        MIRO & 56.29 & 41.28 & 48.79 & 47M \\
        FSDG & 61.77 & 48.43 & 55.10 & 49M \\
        \hline
        \textit{ViT-Based:} & \\
        SDViT & 58.43 & 46.48 & 52.46 & 23M \\
        GMoE & 60.70 & 48.13 & 54.42 & 34M \\
        \hline
       \textit{VMamba-Based:} & \\
        VMamba & 65.01 & 61.93 & 63.47 & 33M \\
        \textbf{HSSH} (Ours) & \textbf{67.48} & \textbf{64.57} & \textbf{66.03} & 35M \\
        \hline
	\end{tabular}
        }
        \caption{Performance comparison of the proposed HSSH and existing methods on the Cub-Paintings dataset.*: Target domain is available when training. }
	\label{tab:cubp}
\end{table}

\begin{table}[!h]
	\centering
	\resizebox{\linewidth}{!}{
	\begin{tabular}{c|ccc|c}
		\hline
		Method &  M $\rightarrow$ N &  N $\rightarrow$ M & Avg & Params\\ 
        \cline{1-5}
        \textit{ResNet-Based:} & \\
        DP-Net* (DA) & 59.27 & 62.01 & 60.64 & 103M \\
        PAN* (DA) & 59.42 & 62.15 & 60.79 & 103M \\
        ARM      &  47.84 &  29.62 &  38.73 & 24M \\
        DANN     &  56.01 &  35.37 &  45.69 & 24M \\
        MLDG     &  56.92 &  35.95 &  46.44 & 23M \\
        GroupDRO &  52.14 &  34.06 &  43.10 & 23M \\
        CORAL    &  54.47 &  33.90 &  44.18 & 23M \\
        SagNet   &  57.91 &  32.58 &  45.24 & 24M \\
        MixStyle &  53.74 &  27.45 &  40.60 & 23M \\
        Mixup    &  50.91 &  36.13 &  43.52 & 23M \\
        RIDG     &  51.95 &  43.88 &  47.92 & 24M \\
        SAGM     &  56.82 &  39.59 &  48.20 & 23M \\
        MIRO     &  55.82 &  39.86 &  47.84 & 47M \\
        FSDG & 57.35 & 50.68 & 54.02 & 49M \\
        \hline
        \textit{ViT-Based:} \\
        SDViT & 54.57 & 48.76 & 51.67 & 23M \\
        GMoE & 56.72 & 50.61 & 53.67 & 34M\\
        \hline
        \textit{VMamba-Based:} & \\
        Vmamba & 60.14 & 73.55 & 66.85 & 33M\\
        \textbf{HSSH} (Ours) & \textbf{62.60} & \textbf{76.69} & \textbf{69.65} & 35M\\
        \hline
	\end{tabular}
        }
        \caption{Performance comparison of the proposed HSSH and existing methods on the remote sensing dataset. *: Target domain is available when training.}
	\label{tab:rs}
\end{table}

\begin{table*}[!t]
	\centering
	\resizebox{0.80\linewidth}{!}{
	\begin{tabular}{c|ccccccc|c}
		\hline
		Method & C $\rightarrow$ I &  C $\rightarrow$ N &  I $\rightarrow$ C &  I $\rightarrow$ N &  N $\rightarrow$ C &  N $\rightarrow$ I & Avg & Params\\ 
        \cline{1-9}
        \textit{ResNet-Based}: & \\
        DP-Net* (DA) & 68.37 & 82.66 & 90.20 & 87.52 & 91.47 & 75.13 & 82.56 & 103M \\
        PAN* (DA) & 69.79 & 84.19 & 90.46 & 88.10 & 92.51 & 75.03 & 83.34 & 103M \\
        ARM & 50.51 & 71.25 & 77.38 & 74.20 & 84.74 & 59.82 & 69.65 & 24M \\
        DANN & 52.75 & 71.82 & 80.79 & 73.59 & 85.55 & 61.53 & 71.01 & 24M \\
        MLDG & 53.55 & 72.19 & 80.74 & 74.83 & 85.61 & 61.95 & 71.48 & 23M \\
        GroupDRO & 52.61 & 70.78 & 81.87 & 74.40 & 86.26 & 61.32 & 71.21 & 23M \\
        CORAL & 54.64 & 72.93 & 81.01 & 74.97 & 86.10 & 62.51 & 72.03 & 23M \\
        SagNet & 53.66 & 71.75 & 81.39 & 74.13 & 85.66 & 62.06 & 71.44 & 24M \\
        MixStyle & 49.95 & 69.04 & 74.46 & 6834 & 83.60 & 57.12 & 67.09 & 23M \\
        Mixup & 52.36 & 71.65 & 82.36 & 75.17 & 85.61 & 62.34 & 71.58 & 23M \\
        RIDG & 47.15 & 66.71 & 82.47 & 73.63 & 85.77 & 60.98 & 69.45 & 24M \\
        SAGM & 54.04 & 73.63 & 82.96 & 77.01 & 87.88 & 63.49 & 73.17 & 23M \\
        MIRO & 54.39 & 74.87 & 82.36 & 75.34 & 86.42 & 62.48 & 72.64 & 47M \\
        FSDG & 66.70 & 83.29 & 89.97 & 87.33 & 92.20 & 74.34 & 82.31 & 49M \\
        \hline
        \textit{ViT-Based:} & \\
        SDViT & 62.21 & 79.91 & 86.58 & 82.87 & 88.26 & 69.64 & 78.25 & 23M \\ 
        GMoE & 64.86 & 82.45 & 89.07 & 85.76 & 90.92 & 73.59 & 81.11 & 34M \\ 
        \hline
        \textit{VMamba-Based:} & \\
        VMamaba & 77.98 & 87.05 & 92.76 & 92.00 & 93.86 & 85.80 & 88.24 & 33M \\
        \textbf{HSSH} (Ours) & \textbf{80.43} & \textbf{90.39} & \textbf{95.25} & \textbf{94.33} & \textbf{96.17} & \textbf{87.57} & \textbf{90.69} & 35M \\
        \hline
	\end{tabular}
        }
        \caption{Performance comparison of the proposed HSSH and existing methods on the Brids-31 dataset. *: Target domain is available when training. }
	\label{tab:bird}
\end{table*}

\begin{table}[!t]
\begin{tabular}{ccc|ccc}
\hline
Backbone & SSH & HMC & $\mathrm{C} \rightarrow \mathrm{P}$ & $\mathrm{P} \rightarrow \mathrm{C}$ & Avg \\
\hline
$\checkmark$ &  &  & 65.01 & 61.93 & 63.47 \\
$\checkmark$ & $\checkmark$ &  & 66.42 & 63.29 & 64.86 \\
$\checkmark$ & $\checkmark$ & $\checkmark$ & \textbf{67.48} & \textbf{64.57} & \textbf{66.03} \\
\hline
\end{tabular}
\caption{Ablation study on each component of the proposed HSSH. Classification accuracy presented in percentage (\%).}
\label{tab:abl}
\end{table}

\begin{table}[h]
\centering
\begin{tabular}{cccc|ccc}
\hline
$\boldsymbol{F}^{1}$ & $\boldsymbol{F}^{2}$ & $\boldsymbol{F}^{3}$ & $\boldsymbol{F}^{4}$ & $\mathrm{C} \rightarrow \mathrm{P}$ & $\mathrm{P} \rightarrow \mathrm{C}$ & Avg \\ 
\cline{1-7}
\hline
$\checkmark$ & ~ & ~ & ~ & 64.78 & 61.96 & 63.53 \\
$\checkmark$ & $\checkmark$ & ~ & ~ & 66.02 & 62.37 & 64.20 \\
$\checkmark$ & $\checkmark$ & $\checkmark$ & ~ & 66.81 & 63.28 & 65.05 \\
$\checkmark$ & $\checkmark$ & $\checkmark$ & $\checkmark$ & \textbf{67.48} & \textbf{64.57} & \textbf{66.03} \\
\hline
\end{tabular}
\caption{Ablation studies on each stage of SSH. Classification accuracy presented in percentage (\%).}
\label{table:abl2}
\end{table}

\begin{figure}[!t]
    \centering
    \includegraphics[width=1.0\linewidth]{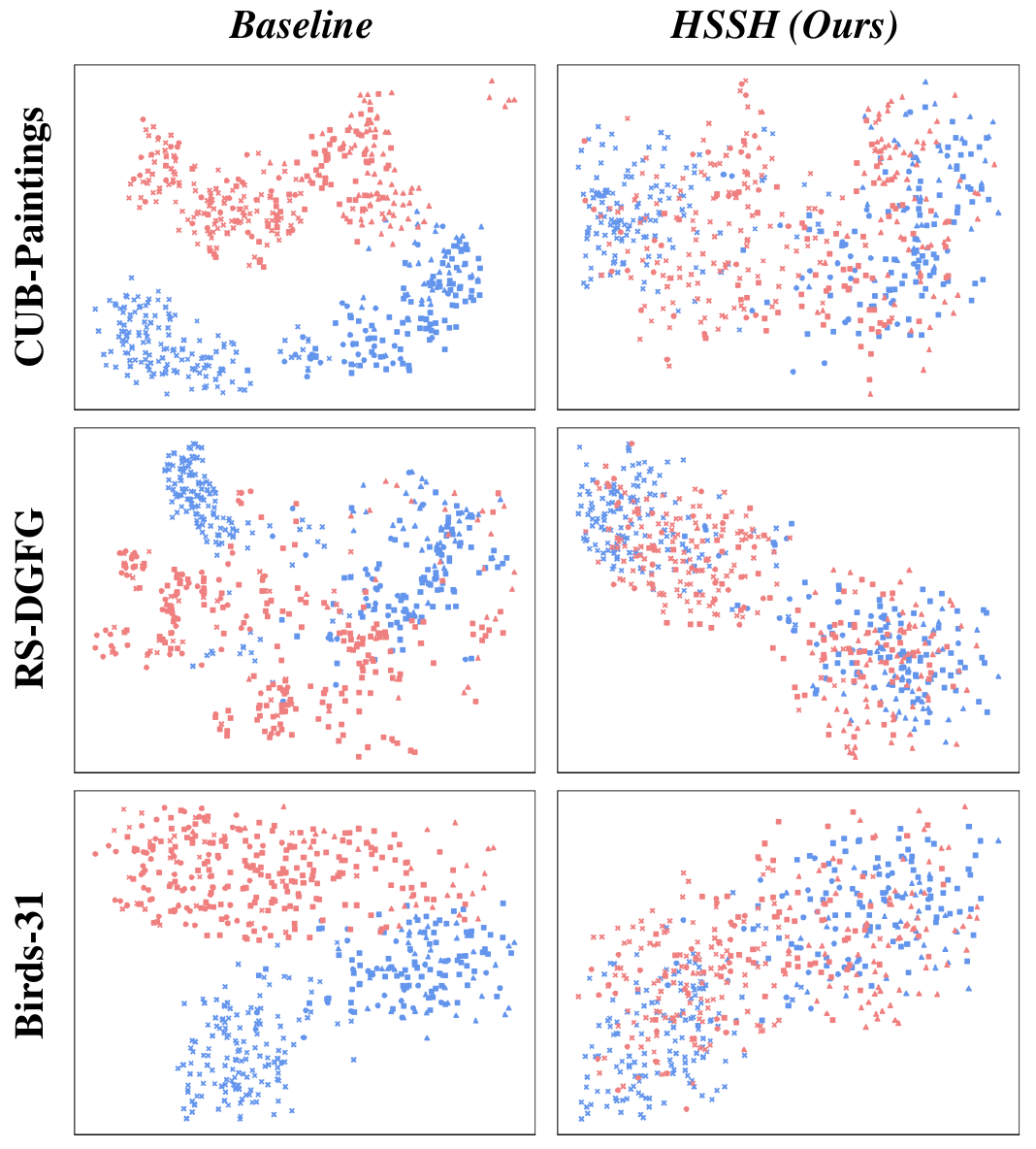}
    \caption{t-SNE feature visualization. Blue: source domain; Red: unseen target domain. Different types of icon refer to different fine-grained categories.}
    \label{fig:tsne}
\end{figure}

\begin{figure*}[!t]
    \centering
    \includegraphics[width=1.0\linewidth]{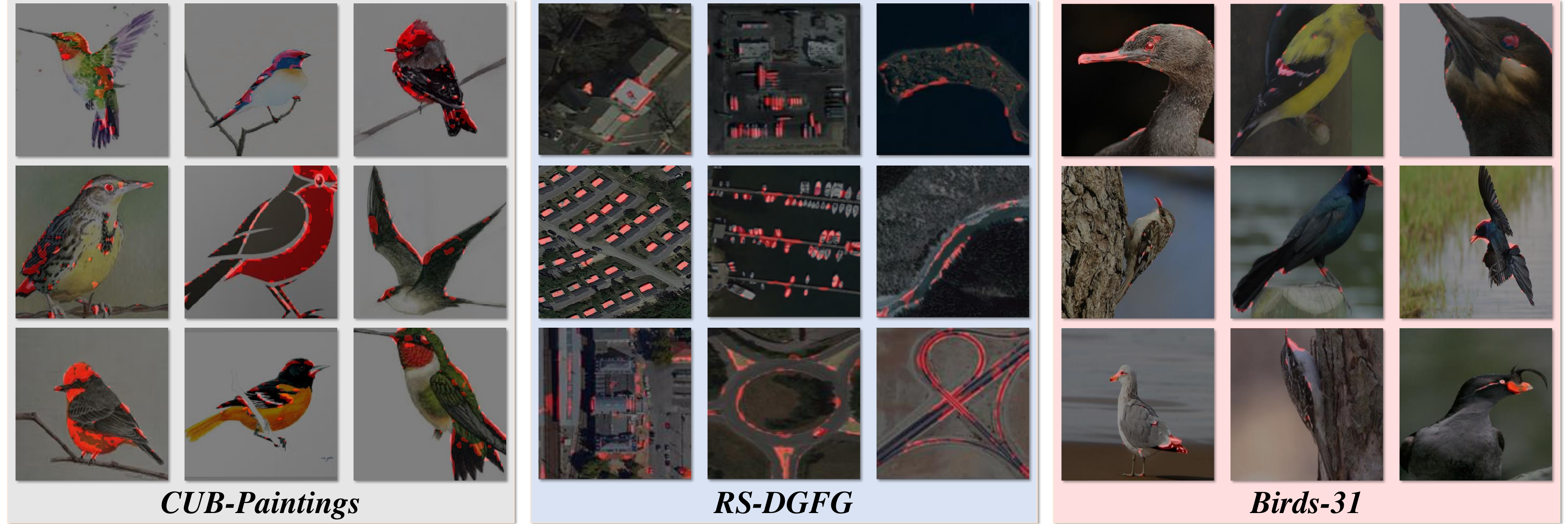}
    \vspace{-0.5cm}
    \caption{Heatmap visualization of the proposed HSSH on unseen target domains by GradCAM. }
    \label{fig:vis}
\end{figure*}

\subsection{Comparison with SOTA}

The compared methods include: 1) generic domain generalization methods that are not especially devised for fine-grained images, namely, ARM \cite{zhang2020adaptive}, DANN \cite{ganin2016domain}, MLDG \cite{li2018learning}, GroupDRO \cite{sagawa2019distributionally}, CORAL \cite{sun2016deep}, SagNet \cite{nam2021reducing}, MixStyle \cite{zhou2021domain}, Mixup \cite{yan2020improve}, RIDG \cite{chen2023domain}, SAGM \cite{wang2023sharpness}, MIRO \cite{cha2022domain}, SDViT \cite{sultana2022self}, and GMoE \cite{li2023sparse}; 
2) domain adaptation fine-grained methods, namely, PAN \cite{du2021progressive} and DP-Net \cite{wang2021dynamic};
3) domain generalized fne-grained methods FSDG \cite{yu2024fine}.

\noindent \textbf{Results on CUB-Paintings} are presented in Table~\ref{tab:cubp}. The proposed HSSH outperforms all other approaches, even surpassing domain adaptation (DA) methods that have access to target domain images. 
Compared to the second-best FGDG method FSDG, HSSH shows an accuracy improvement of 5.71\% and 16.14\% in the C$\rightarrow$P and P$\rightarrow$C settings. 
Additionally, by expanding the style in the state space and hierarchically alignment in hyperbolic space, HSSH further enhances VMamba's performance to achieve improvements of 2.47\% and 2.64\% in the two respective settings.

\noindent \textbf{Results on RS-FGDG} are presented in Table ~\ref{tab:rs}. The remote sensing fine-grained classification task is more challenging compared to CUB-Paintings due to the complexity of spatial distributions and varying resolutions, leading to a noticeable limited performance for DG methods. Despite this, VMamba achieves the second-best performance in DG method, outperforming FSDG by 5.25\% and 26.01\% in the M$\rightarrow$N and N$\rightarrow$M settings, respectively. However, VMamba does not surpass the DA methods DP-Net and PAN. The proposed HSSH scheme demonstrates superior performance, achieving improvements of 2.46\% and 3.14\% over VMamba and even outperforming DA methods without access to any target domain when training.

\noindent \textbf{Results on Birds-31} are presented in Table ~\ref{tab:bird}. The experiments were conducted across six different settings spanning three domains. The Birds-31 setup involves three natural image datasets with relatively smaller domain gaps, leading to higher performance for both DG and DA methods. The proposed HSSH achieves state-of-the-art results in all settings, outperforming the best DA method PAN, with improvements of 10.64\%, 6.20\%, 4.79\%, 3.36\%, 12.54\%, and 7.35\% for C$\rightarrow$I, C$\rightarrow$N, I$\rightarrow$C, I$\rightarrow$N, N$\rightarrow$C, and N$\rightarrow$I, respectively. 
Notably, the average accuracy has been improved by 8.38\% compared to the second-best FGDG method FSDG. 

\subsection{Ablation Study}

\noindent \textbf{On Each Component.} 
The proposed HSSH consists of the VMamba backbone, state space hallucination module and hyperbolic manifold consistency, which we denote as backbone, SSH and HMC, respectively. 
For ablation studies, when implementing each component, the corresponding loss function is also implemented.
The results under the CUB-Paintings settings are shown in Table~\ref{tab:abl}. 
Specifically, when SSH is utilized, the classification accuracy improvement is 1.41\%, 1.36\%, and 1.39\% under C$\rightarrow$P, P$\rightarrow$C, and the average setting, respectively, compared to using $\mathcal{L}_{cls}$ alone for FGDG. 
The use of HMC further improves the classification accuracy by 1.06\%, 1.28\%, and 1.17\%, demonstrating the positive impact of hierarchical hyperbolic embedding on fine-grained representation.

\noindent \textbf{On Hallucinated Stage.} 
The hallucinated features $\widetilde{\mathbf{F}}^{1}$, $\widetilde{\mathbf{F}}^{2}$, $\widetilde{\mathbf{F}}^{3}$, and $\widetilde{\mathbf{F}}^{4}$ expand the style space at different stages within the state space. 
An ablation study is conducted on each stage. 
As shown in Table~\ref{table:abl2}, hallucinating across all scales yields optimal performance on the CUB-Paintings. Notably, expanding the style space at the final stage $\mathbf{F}^{4}$ is most critical, resulting in improvements of 0.67\%, 1.29\%, and 0.98\% under the C$\rightarrow$P, P$\rightarrow$C, and average setting, respectively.

\subsection{Visualization}

\noindent \textbf{On Feature Space.}
To inspect if HSSH alleviates the gap between each fine-grained domain over the VMamba baseline, we extract the state embeddings from both the source and unseen target domains.
Then, we display them by t-SNE feature visualization, and the results are shown in Fig.~\ref{fig:tsne}.
For all experiments, blue and red points refer to the samples from the source and unseen target domain, respectively.
HSSH allows a more uniform mixture between the source and target domain samples than the VMamba baseline.
This indicates its effectiveness to handle the problem of FGDG. 

\noindent \textbf{On Activated Patterns.}
Following the visualization of prior FGVC methods, GradCAM is used to visualize the features from the proposed HSSH.
The results are displayed in Fig.~\ref{fig:vis}. 
Overall, the proposed HSSH is able to discern the subtle and tiny fine-grained patterns on the unseen domains from all the three benchmarks, demonstrating its effectiveness. 

\section{Conclusion}

In this paper, we propose a novel Hyperbolic State Space Hallucination (HSSH) scheme for fine-grained domain generalization (FGDG). 
It makes an earlier exploration on not only harnessing state space model (SSM) for FGDG, but also modeling the hyperbolic selective state space. 
The proposed HSSH consists of two key components, namely, state space hallucination (SSH) and hyperbolic manifold consistency (HMC).
SSH enriches the style diversity by firstly extrapolating and then hallucinating the style of the state embeddings.
HMC minimizes the hyperbolic distance between the pre- and post- style hallucinate images, which benefits the discernment of fine-grained patterns despite the cross-domain styles.
Finally, HMC minimizes the hyperbolic distance, so that the impact from the style variation on fine-grained patterns can be eliminated.
Extensive experiments across three FGDG benchmarks demonstrate the state-of-the-art performance of the proposed HSSH.


\bibliography{aaai25}

\end{document}